\documentclass[11pt,a4paper]{article}
\usepackage[hyperref]{ACL/acl2020}
\usepackage{times}
\usepackage{latexsym}

\usepackage{microtype}

\aclfinalcopy % Uncomment this line for the final submission
\usepackage{latexsym}
\usepackage{booktabs} % For formal tables
\usepackage{amsmath}
\usepackage{amsfonts}
\usepackage{bm}
\usepackage{array}
\usepackage{enumitem}
\usepackage{needspace}
\usepackage{adjustbox}
\usepackage{tikz}
\usepackage{pgfplots}
\usepackage{multirow}
\usepackage{autobreak}
\usepackage{makecell}
\usepackage{xcolor}
\usepackage{amssymb}
\usepackage{import}
\usepackage{subcaption}
\usepackage{nicefrac}       % compact symbols for 1/2, etc.
\usepackage{microtype}      % microtypography
\usepackage{tabularx}
\usepackage{dcolumn}
\usepackage{algorithm}
\usepackage{algorithmic}
\usepackage{stmaryrd}
\newcommand*{\rom}[1]{\romannumeral#1\relax}

\newcommand\setrow[1]{\gdef\rowmac{#1}#1\ignorespaces}
\newcommand\clearrow{\global\let\rowmac\relax}
\newcommand{\down}{}
\clearrow

\title{
    Data Manipulation: Towards Effective Instance Learning\\ for Neural Dialogue Generation via Learning to Augment and Reweight
}

\author{
Hengyi Cai$^{\dagger,\S}$\thanks{\ \ Work done at Data Science Lab, JD.com.}, Hongshen Chen$^\ddagger$\\
{\bf Yonghao Song$^\dagger$, Cheng Zhang$^\dagger$, Xiaofang Zhao$^\dagger$ {\normalfont{and}} Dawei Yin$^{\dagger\dagger}$} \\
$^\dagger${Institute of Computing Technology, Chinese Academy of Sciences, Beijing, China} \\
$^\S${University of Chinese Academy of Sciences, Beijing, China} \\
$^\ddagger${Data Science Lab, JD.com, China} \\
$^{\dagger\dagger}${Baidu Inc., China} \\
{caihengyi@ict.ac.cn, ac@chenhongshen.com,} \\
{\{songyonghao, zhangcheng, zhaoxf\}@ict.ac.cn, yindawei@acm.org}
}

\date{}

\begin{document}
\maketitle
\begin{abstract}
Current state-of-the-art neural dialogue models learn from human conversations following the data-driven paradigm.
As such, a reliable training corpus is the crux of building a robust and well-behaved dialogue model.
However, due to the open-ended nature of human conversations, the quality of user-generated training data varies greatly, and effective training samples are typically insufficient while noisy samples frequently appear.
This impedes the learning of those data-driven neural dialogue models.
Therefore, effective dialogue learning requires not only more reliable learning samples, but also fewer noisy samples.
In this paper, we propose a data manipulation framework to proactively reshape the data distribution towards reliable samples by augmenting and highlighting effective learning samples as well as reducing the effect of inefficient samples simultaneously.
In particular, the data manipulation model selectively augments the training samples and assigns an importance weight to each instance to reform the training data. 
Note that, the proposed data manipulation framework is fully data-driven and learnable.
It not only manipulates training samples to optimize the dialogue generation model, but also learns to increase its manipulation skills through gradient descent with validation samples.
Extensive experiments show that our framework can improve the dialogue generation performance with respect to various automatic evaluation metrics and human judgments.
\end{abstract}
\section{Introduction}

Open-domain dialogue generation, due to its potential applications, is becoming ubiquitous in the community of natural language processing.
Current end-to-end neural dialogue generation models~\citep{DBLP:conf/naacl/LiGBGD16,DBLP:journals/corr/SerbanSLCPCB16,DBLP:conf/acl/ZhaoZE17} are primarily built following the data-driven paradigm, that is, these models mimic the human conversations by training on the large-scale query-response pairs.
As such, a reliable training corpus that exhibits high-quality conversations is the crux of building a robust and well-behaved dialogue model.

\begin{figure}[!t]
    \centering
      \includegraphics[width=1.0\columnwidth]{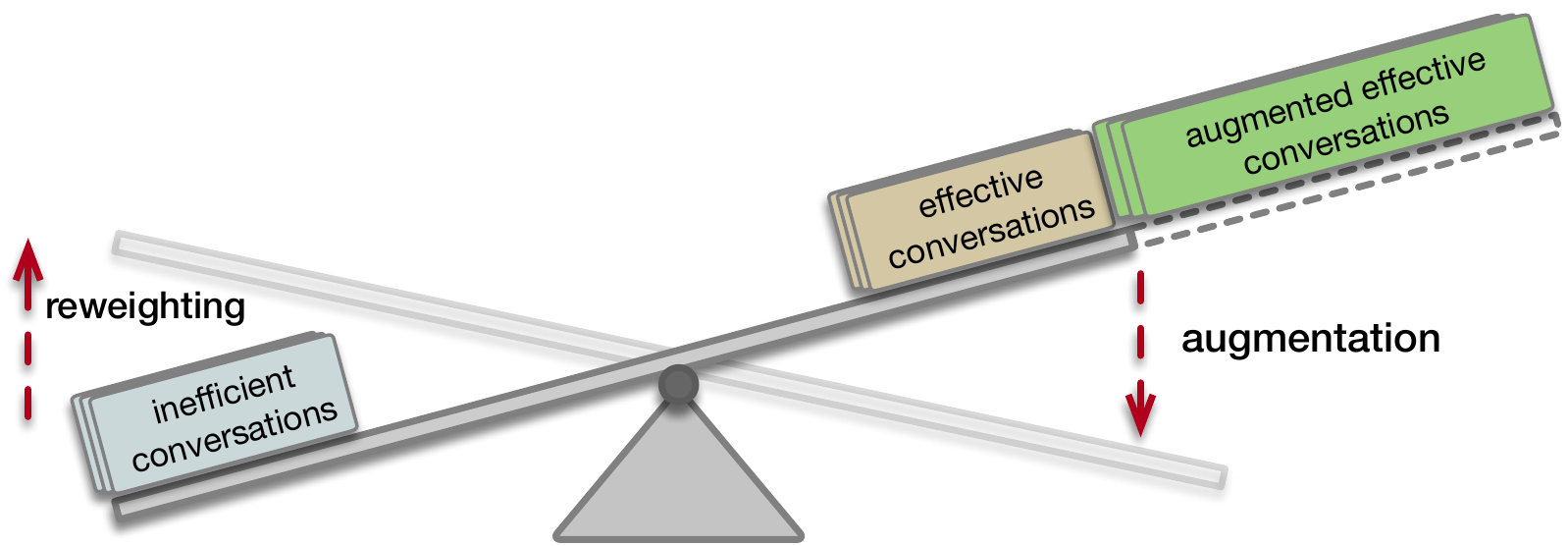}
      \caption{
        Data manipulation helps the dialogue model training by augmenting and highlighting effective learning samples as well as reducing the weights of inefficient samples.
      }
      \label{fig:motivation_balance}
\end{figure}

Unfortunately, owing to the subjectivity and open-ended nature of human conversations, the quality of the collected human-generated dialogues varies greatly~\citep{DBLP:conf/ijcai/ShangFPFZY18}, which hampers the effectiveness of data-driven dialogue models:
1) Effective conversation samples are quite insufficient. 
To glean some insights on the data quality of dialogue corpus, we choose the query-relatedness to take a glimpse of the data quality. 
In dialogue corpus, some conversations are quite coherent, where the queries and responses are well-correlated, while others are not. 
Query-relatedness measures the semantic similarities between the query and its corresponding response in the embedding space and ranges from 0 to 1.  
When reviewing DailyDialog~\citep{DBLP:conf/ijcnlp/LiSSLCN17}, we find that only 12\% conversation samples are of relatively high query-relatedness scores ($>0.6$).
Without adequate reliable training samples, the neural dialogue model is prone to converge to a sub-optimal {point}.
2) Meanwhile, noisy and even meaningless conversation samples frequently appear. 
As \citet{DBLP:conf/naacl/LiGBGD16} reported, ``I don't know'' appears in over 113K sentences in the training corpus OpenSubtitles~\citep{DBLP:conf/lrec/LisonT16}.
Such kind of noisy conversation data prevails in neural dialogue model training, and vitally impedes the model learning.

Therefore, effective dialogue learning requires not only more reliable learning samples, but also fewer noisy samples.
In this work,  as illustrated in Figure~\ref{fig:motivation_balance}, we propose a novel learnable data manipulation framework to  proactively reshape the data distribution towards reliable samples by augmenting and highlighting effective learning samples as well as reducing the weights of inefficient samples simultaneously. 
Specifically, to generate more effective data samples, the data manipulation model selectively augments the training samples in terms of both word level and sentence level, using masked language models such as BERT~\citep{DBLP:conf/naacl/DevlinCLT19} and back-translation~\citep{sennrich-etal-2016-improving} technique. 
To reduce the weights of inefficient samples from the original training samples and the augmented samples, the data manipulation model assigns an importance weight to each sample to adapt the sample effect on dialogue model training.
It gives out higher importance weights to critical learning samples and lower weights to those inefficient samples. 
Furthermore, different from most previous data augmentation or data weighting studies~\citep{Li2019InsufficientDC,DBLP:conf/ijcai/ShangFPFZY18,csaky-etal-2019-improving}, which are unaware of the target model states during augmentation or weighting, our data manipulation framework not only manipulates training samples to optimize the dialogue generation model, but also learns to increase its manipulation skills through gradient descent with validation samples.

We apply the proposed data manipulation framework on several state-of-the-art generation models with two real-life open-domain conversation datasets and compare with the recent data manipulation approaches in terms of 13 automatic evaluation metrics and human judgment.
{Experiment results show that our data manipulation framework outperforms the baseline models over most of the metrics on both datasets.}

\section{Data Manipulation for Neural Dialogue Generation}
\begin{figure}[!t]
\centering
  \includegraphics[width=0.99\columnwidth]{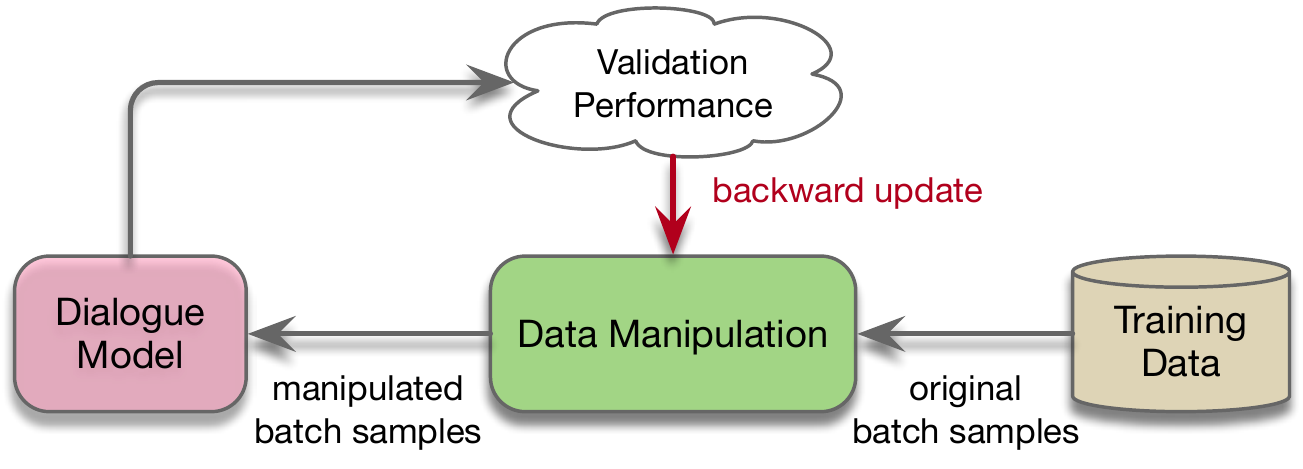}
  \caption{
    Overview of the proposed automated data manipulation framework for neural dialogue generation.
    At training step $t$, the data manipulation model augments and weights the training samples for the dialogue model learning.
  }
  \label{fig:model_arch}
\end{figure}

 The proposed data manipulation framework tackles the problem of un-even quality data by inducing the model learning from more effective dialogue samples and reducing effects of those inefficient samples simultaneously.
 In particular, as illustrated in Figure~\ref{fig:model_arch}, it manipulates and reshapes the data distribution for neural dialogue model learning in mainly three stages:
 First, each batch of training samples are selectively augmented to generate more variant samples;
 and then, all the samples, including the original samples and the augmented samples, are assigned with instance weights indicating their importance regarding current learning status;
 finally, the weighted samples are fed into the neural dialogue model to induce the model learning from more effective training instances. 
 
 Note that, although we describe the framework in three components for ease of understanding, in fact, the whole framework can be trained in an end-to-end manner.
 As a result, the data manipulation network is capable of not only manipulating training samples to optimize the dialogue generation model, but also learning to increase its manipulation skills through gradient descent with validation samples.
 
 We first introduce the augmentation and weighting strategies for data manipulation in \S\ref{augmentation} and \S\ref{weighting}, and then describe how the neural dialogue generation model learns from the manipulated samples in \S\ref{sec:gen_with_manipulation}.
 Parameters estimation for the data manipulation model is elaborated in \S\ref{manipulation_learning}.

\begin{figure*}[!t]
  \centering
    \includegraphics[width=0.98\textwidth]{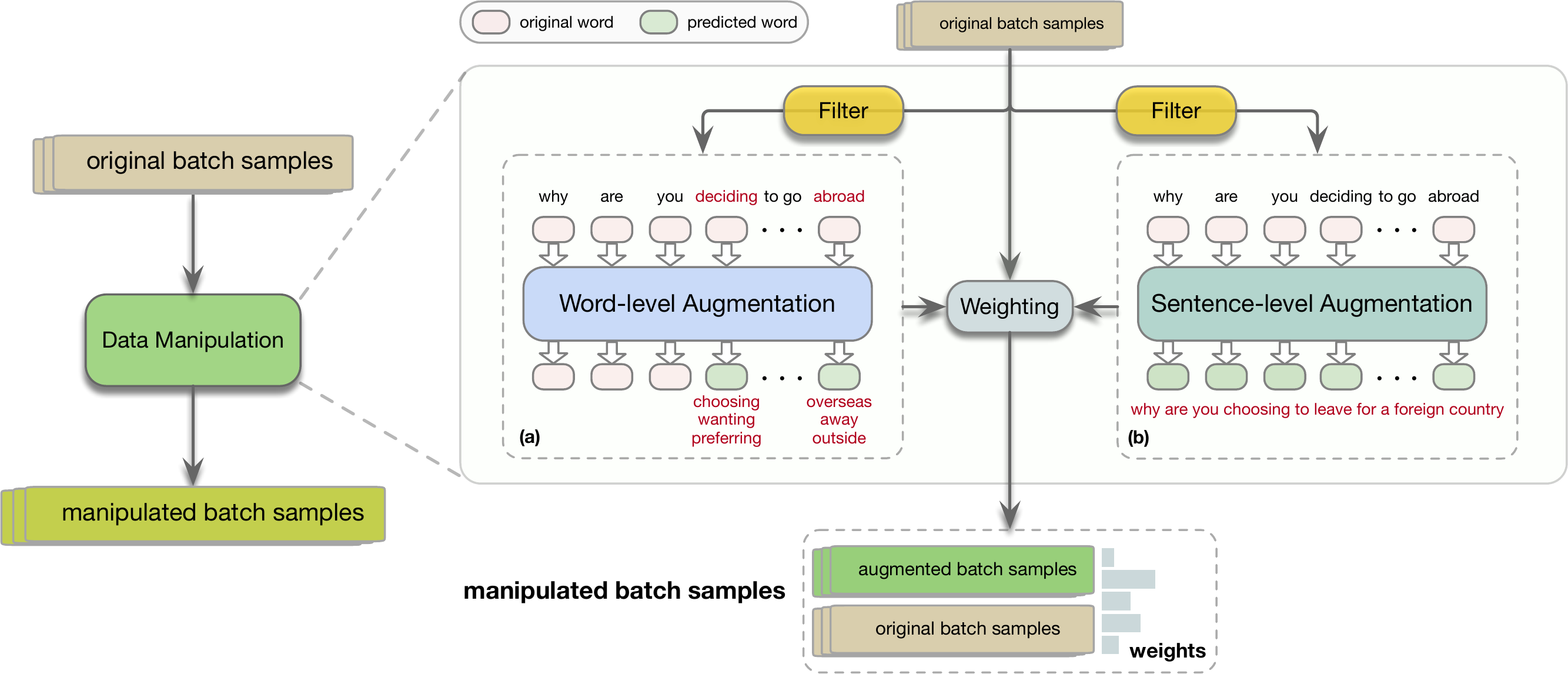}
    \caption{
      Illustration of the data manipulation model.
      During training, it takes the original batch samples as input, and generates the augmented data samples as well as the importance weights for dialogue model training.
    }
    \label{fig:data_manipulation_net}
\end{figure*}

\subsection{Dialogue Augmentation}
\label{augmentation}
 To induce the neural dialogue generation model to learn from more effective samples, we develop a gated data augmentation mechanism for the manipulation framework to selectively augment the learning samples.
 
 Specifically, as shown in  Figure~\ref{fig:data_manipulation_net},
 given a training sample, the manipulation framework first specifies whether to augment it or not through {an instance filter, which can be implemented using a $sigmoid$ gating function.}
 Then, two levels of data augmentation are introduced, word-level contextual augmentation and sentence-level data augmentation, to augment the chosen sample accordingly.

\subsubsection{Word-level Contextual Augmentation}
 As the name suggests, word-level augmentation enriches the training samples by substituting the words in the original sample (Figure~\ref{fig:data_manipulation_net} (a)).
 Here, we employ a masked language model, BERT~\citep{DBLP:conf/naacl/DevlinCLT19}, to implement word-level augmentation.
 Given an original sentence\iffalse $\hat{x}$\fi, the language model first randomly masks out a few words. 
 BERT then takes in the masked sentence and predicts the corresponding masked positions with new words.

 A fixed pre-trained BERT may not generalize well for our data manipulation framework, because BERT is unaware of the dialogue learning status.
 To mitigate such defects, we further fine-tune BERT through backpropagation \iffalse to minimize the validation loss\fi {(more details in \S~\ref{manipulation_learning}).}
 In particular, BERT is adapted to be differentiable by utilizing a gumbel-softmax approximation~\citep{DBLP:conf/iclr/JangGP17} when predicting substitution words.

\subsubsection{Sentence-level Data Augmentation}
Word-level data augmentation is quite straightforward. 
However, such kind of rewriting is limited to only a few words. 
In human dialogues, there exist various synonymous conversations with different sentence structures.
To further diversify the expressions in conversion, we introduce the sentence-level data augmentation through back-translation as in \citet{DBLP:conf/emnlp/EdunovOAG18,DBLP:conf/iclr/YuDLZ00L18}, which trains two translation models: one translation model from the source language to target language and another backward translation model from the target to the source, as shown in Figure~\ref{fig:data_manipulation_net} (b).
By transforming the expression styles across different languages, the augmented training samples are expected to convey similar information while with different expressions.

Similar to the fine-tuning strategy in word-level data augmentation, we also fine-tune the sentence-level data augmentation components to encourage the model to generate more effective samples for dialogue training.
The gradients are back-propagated into the translation-based augmentation model, where a differentiable gumbel-softmax is utilized when predicting sentences using the translation model.

\subsection{Data Weighting}
\label{weighting}
Given the original training samples and the augmented samples, to deal with the problem of noisy instances, data manipulation model assigns an importance weight to each training sample regarding the learning status.
In particular, the sample importance weights are approximated through a softmax function over the scores of these instances.
A multilayer perceptron is employed to compute example scores, taking distributional representations of these instances as input.
Each sample is converted into its corresponding  distributional representation through a transformer-based encoder.

\subsection{Dialogue Generation with Data Manipulation}
\label{sec:gen_with_manipulation}
Conventionally, neural dialogue generation model is optimized with a vanilla negative log-likelihood loss using the training data $\mathcal{D}$ with size $N$: $\mathcal{L}_{vanilla}=\sum_{j=1}^{N}{-\log{p({\bm{y}_j}|{\bm{x}_j})}}$,
where each sample is treated equally.
In our framework, we assign each sample with an importance weight {and augment the original training set $\mathcal{D}=\{(\bm{x}_j, \bm{y}_j)\}_{j=1}^{N}$ to $\mathcal{D}'=\{(\bm{x}_j, \bm{y}_j)\}_{j=1}^{N'}$} regarding the learning status.
To perform the weighted optimization {with augmented training set $\mathcal{D}'$}, we utilize a weighted negative log-likelihood loss function:
\begin{equation}
  \begin{split}
    \mathcal{L}_{dm} = \sum_{j=1}^{N'}{-w_j\log{p({\bm{y}_j}|{\bm{x}_j})}},
  \end{split}
  \label{eq:loss_dm}
\end{equation}
where $w_j$ is the instance weight produced by the data manipulation network.

\subsection{Parameter Estimation for Data Manipulation}
\label{manipulation_learning}
 The data manipulation network not only manipulates training samples to optimize the dialogue learning process, but also learns to increase its manipulation skills through gradient descent with validation samples.
 We formulate such joint learning process following a novel policy learning paradigm~\citep{DBLP:journals/corr/abs-1910-12795,DBLP:conf/iclr/TanHYSX19},
 where the manipulation framework is formulated as a learnable data-dependent reward function $R_\phi(\bm{d}=\{\bm{x}, \bm{y}\}|\mathcal{D})$, the dialogue model $p_\theta(\bm{y}|\bm{x})$ is treated as a policy, the input $\bm{x}$ as the ``state'', and the output $\bm{y}$ as the ``action''.
 The reward function $R_\phi(\bm{d}|\mathcal{D})$ is defined as:
\begin{equation}
  \small
  \begin{split}
  R_\phi(\bm{d}|\mathcal{D}) = \left\{
    \begin{array}{ll}
    w_i & \text{if}\ \  \bm{d} \text{ is an augmented sample}\\  
    & \text{of}\ \bm{d}_i^* \ \text{or}\ \bm{d}=\bm{d}_i^*,\  \bm{d}_i^* \in \mathcal{D} \\
    -\infty & \text{otherwise},
    \end{array}
    \right.
  \end{split}
  \label{eq:r-data_mani}
\end{equation}
 where ${\phi}$ denotes the parameter of data manipulation network and $w_i\in \mathbb{R}$ is the importance weight associated with the $i$th data sample.
 In such formulation, a sample $\bm{d}$ receives a real-valued reward when $\bm{d}$ is an augmented sample, or $\bm{d}$ matches an instance in the original training set.

 As depicted in  Algorithm~\ref{algo:optimization}, the parameter $\bm{\theta}$ of the neural dialogue model and parameter $\bm{\phi}$ of the data manipulation network are alternatively optimized.
 {Jointly optimizing the dialogue model and the manipulation network can be regarded as reward learning,} 
 where the policy $p_\theta(\bm{y}|\bm{x})$ receives relatively higher rewards for effective samples and lower rewards for those inefficient samples.
 More concretely, to optimize the neural dialogue model, at each iteration, mini-batch instances are sampled from the training set, and are then enriched through augmentation and weighting.
 The parameter ${\theta}$ of the neural dialogue model is then updated with a weighted negative log-likelihood loss function in Eq.(\ref{eq:loss_dm}):
 \begin{equation}
 {\theta^{'}}={\theta}-\alpha \nabla_{\theta}\mathcal{L}_{dm}({\theta}, {\phi}),
 \label{eq:op_theta}
 \end{equation}
 where $\nabla_{\theta}\mathcal{L}_{dm}({\theta}, {\phi})$ is the gradient of ${\theta}$ with respect to the loss $\mathcal{L}_{dm}$, and $\alpha$ is the step size.
 The parameter ${\phi}$ of the data manipulation network is learned by taking a meta gradient descent step on validation samples~\citep{DBLP:conf/icml/RenZYU18}. 
 Equation (\ref{eq:op_theta}) shows that ${\theta^{'}}$ depends on ${\phi}$. Therefore, the manipulation model (i.e. the reward function $R_\phi(\bm{d}|\mathcal{D})$) can be optimized by directly backpropagating the gradient through ${\theta^{'}}$ to ${\phi}$.

\begin{algorithm}[!t]
    \small
    \centering
    \caption{Joint Learning of Dialogue Model and Data Manipulation Network}
    \label{algo:optimization}
    \begin{algorithmic}[1]
    \REQUIRE The dialogue model $\theta$, data manipulation network $\phi$, training set $\mathcal{D}$ and validation set $\mathcal{D}^v$ 
    \STATE Initialize dialogue model parameter ${\theta}$ and data manipulation model parameter ${\phi}$
    \REPEAT
        \STATE Optimize ${\theta}$ on $\mathcal{D}$ enriched with data manipulation. 
        \STATE Optimize ${\phi}$ by maximizing data log-likelihood on $\mathcal{D}^v$.
    \UNTIL{convergence}
    \ENSURE Learned dialogue model $\theta^{*}$ and data manipulation model $\phi^*$
    \end{algorithmic}
\end{algorithm}
\section{Experiments}
\label{sec:exp}

\begin{table}[!th]
\centering
\resizebox{0.85\columnwidth}{!}{
\begin{tabular}{lccc}
\toprule
Dataset       & Train   & Valid  & Test   \\
\midrule
DailyDialog   & 54,889  & 6,005  & 5,700  \\
OpenSubtitles & 64,000 & 8,000 & 8,000 \\
\bottomrule
\end{tabular}
}
\caption{Data statistics of the experiment corpora.}
\label{tbl:datasets}
\end{table}

\subsection{Experiment Setup}
\paragraph{Data}
We conduct experiments on two English conversation datasets: 
(1) \textit{DailyDialog}~\citep{DBLP:conf/ijcnlp/LiSSLCN17}, a collection of real-world dialogues widely used in open-domain dialogue generation.
This is a multi-turn dataset, and we treat each turn as a training pair in this work.
The overlapping pairs are removed from the data set.
(2) \textit{OpenSubtitles}~\citep{DBLP:conf/lrec/LisonT16}, a group of human-human conversations converted from movie transcripts. 
{80,000 instances are sampled from the original corpus and }the data proportion for train/valid/test set is set to 8/1/1, respectively. 
The dataset statistics are listed in Table~\ref{tbl:datasets}. 

\paragraph{Experimental Models}
To ascertain the effectiveness and applicability of our method, 
we implement the proposed data manipulation framework on following representative models:
(\rom{1}) \textbf{SEQ2SEQ}: a RNN-based sequence-to-sequence model with attention mechanisms~\citep{Bahdanau2014NeuralMT};
(\rom{2}) \textbf{CVAE}:  a latent variable model using conditional variational auto-encoder, trained with KL-annealing and a BoW loss as in ~\citet{DBLP:conf/acl/ZhaoZE17};
(\rom{3}) \textbf{Transformer}:  an encoder-decoder architecture relying solely on the attention mechanisms~\citep{DBLP:conf/nips/VaswaniSPUJGKP17}.

\paragraph{Comparison Models}
We also compare our approach with previous data augmentation or instance weighting methods:
(\rom{1}) \textbf{CVAE-GAN}~\cite{Li2019InsufficientDC}:  a model that combines CVAE and GAN for augmenting the training data to generate more diversified expressions.
(\rom{2}) \textbf{Calibration}~\cite{DBLP:conf/ijcai/ShangFPFZY18}: a calibration network measures the quality of data samples and enables weighted training for dialogue generation.
(\rom{3}) \textbf{Clustering}~\citep{csaky-etal-2019-improving}: it clusters high-entropy samples as noises and filters them out.

\begin{table*}[!t]
\centering
\resizebox{1.0\textwidth}{!}{
\centering
\begin{tabular}{p{0.05\columnwidth}|>{\rowmac}l>{\rowmac}c>{\rowmac}c>{\rowmac}c>{\rowmac}c>{\rowmac}c>{\rowmac}c>{\rowmac}c>{\rowmac}c>{\rowmac}c>{\rowmac}c>{\rowmac}c>{\rowmac}c>{\rowmac}c<{\clearrow}}
\toprule
 & Models & Dist-1 & Dist-2 & Dist-3 & Intra-1 & Intra-2 & Intra-3 & Ent-1 & Ent-2 & Ent-3 & BLEU & Avg & Ext & Gre \\
\midrule
\multirow{6}{*}{(a)} & SEQ2SEQ & 0.9026 & 4.2497 & 8.4039 & 87.909 & \textbf{94.399} & 95.971 & 6.7263 & 10.381 & 12.036& 0.2160 & 67.671 & 47.472 & 68.349 \\ 
\setrow{\bfseries} 
 &\normalfont{SEQ2SEQ} (${\bigstar}$) & 1.3058 & 5.8408 & 11.2820 & 88.628 & \normalfont{94.268} & 96.171 & 7.0253 & 11.018 & 12.726 & 0.3619 & 68.018 & 47.665 & 68.708   \\ \cline{2-15} 
 &CVAE & 0.9798 & 4.6095 & 9.0876 & 91.848 & 96.815 & 98.025 & 6.9184 & 10.740 & 12.365 & 0.2617 & \textbf{66.935} & 46.926 & 68.068  \\  
\setrow{\bfseries} 
 &\normalfont{CVAE} (${\bigstar}$)    & 2.0683 & 9.0082 & 17.3260 & 93.301 & 97.418 & 98.323 & 7.0278 & 11.078 & 12.586 & 0.2954 & \normalfont{66.363} & 46.955 & 68.424  \\ \cline{2-15} 
 &Transformer & 1.3489 & 5.9736 & 11.3310 & 87.725 & 94.170 & 95.944 & 6.9024 & 10.624 & 11.941 & 0.2342 & 65.305 & 46.223 & 67.419  \\ 
\setrow{\bfseries} 
 &\normalfont{Transformer} (${\bigstar}$) & 2.4763 & 11.6270 & 21.4520 & 89.058 & 96.615 & 98.248 & 7.1556 & 11.320 & 12.956 & 0.4163 & 66.908 & 46.284 & 67.656  \\
\bottomrule
\bottomrule
\multirow{6}{*}{(b)} & SEQ2SEQ 
& 0.5695 & 2.9952 & 6.2377 & \textbf{96.200} & 97.754 & 98.355 & 6.5996 & 10.371 & 12.213 & 0.0078 & 55.912 & 40.320 & 57.664 \\  
\setrow{\bfseries} 
 &\normalfont{SEQ2SEQ} (${\bigstar}$)     & 0.7285 & 3.6053 & 7.2580 & \normalfont{95.938} & 97.829 & 98.561 & {6.8391} & 10.903 & 13.411 & 0.0210 & {58.105} & 41.113 & 59.551   \\ \cline{2-15} 
 &CVAE 
 & 0.5493 & 2.9585 & 6.3159 & 78.534 & 90.028 & \textbf{98.864} & 5.8675 & 10.089 & \textbf{12.544} & 0.0019 & 54.508 & 41.262 & \textbf{62.139} \\ 
\setrow{\bfseries} 
 &\normalfont{CVAE} (${\bigstar}$)        
 & 1.0883 & 4.8967 & 9.7060 & 95.489 & 97.579 & \normalfont{98.201} & 6.8952 & 10.902 & \normalfont{12.200} & 0.0173 & 56.473 & 41.678 & \normalfont{59.330}  \\ \cline{2-15} 
 &Transformer & 0.7226 & 3.8053 & 8.3877 & 92.94 & 94.947 & 96.023 & 7.0361 & 11.091 & 11.832 & 0.0050 & \textbf{55.257} & \textbf{41.302} & 58.232 \\ 
\setrow{\bfseries}
 &\normalfont{Transformer} (${\bigstar}$) & 1.7264 & 6.8750 & 12.5770 & 94.223 & 97.204 & 98.055 & 7.0493 & 11.334 & 12.098 & 0.0110 & \normalfont{55.219} & \normalfont{40.701} & 59.081 \\ 
\bottomrule
\end{tabular}
}
\caption{Automatic evaluation results (\%) on (a) DailyDialog and (b) OpenSubtitles. 
``${\bigstar}$'' denotes that the model is trained using our proposed data manipulation framework.
{The metrics Average, Extrema and Greedy are abbreviated as Avg, Ext and Gre, respectively.}
The best results in each group are highlighted with \textbf{bold}.
}
\label{tbl:main_res}
\end{table*}
\begin{table*}[ht]
    \centering
    \resizebox{1.0\textwidth}{!}{
    \centering
    \begin{tabular}{p{0.05\columnwidth}|>{\rowmac}l>{\rowmac}c>{\rowmac}c>{\rowmac}c>{\rowmac}c>{\rowmac}c>{\rowmac}c>{\rowmac}c>{\rowmac}c>{\rowmac}c>{\rowmac}c>{\rowmac}c>{\rowmac}>{\rowmac}c>{\rowmac}c<{\clearrow}}
    \toprule
     & Models & Dist-1 & Dist-2 & Dist-3 & Intra-1 & Intra-2 & Intra-3 & Ent-1 & Ent-2 & Ent-3 & BLEU & Avg & Ext & Gre  \\ 
    \midrule
    \multirow{4}{*}{(a)} & {Calibration}~\citep{DBLP:conf/ijcai/ShangFPFZY18} & 0.7278 & 3.2265 & 6.0570 & 86.619 & 91.697 & 93.753 & 6.7827 & 10.439 & 11.867 & 0.1876 & 67.309 & 47.347 & 67.886   \\ 
     & {CVAE-GAN}~\citep{Li2019InsufficientDC} & 0.6996 & 3.2448 & 6.4911 & 85.329 & 92.804 & 94.953 & 6.8184 & 10.425 & 12.260 & 0.2149 & 68.012 & 47.079 & 68.007    \\ 
     & {Clustering}~\citep{csaky-etal-2019-improving} & 0.6532 & 3.0747 & 6.2315 & 78.612 & 87.268 & 91.151 & 6.8554 & 10.436 & 12.358 & 0.2062 & \textbf{69.040} & 47.367 & 68.276  \\  
     \setrow{\bfseries} 
     & Ours & 1.3058 & 5.8408 & 11.2820 & 88.628 & 94.268 & 96.171 & 7.0253 & 11.018 & 12.726 & 0.3619 & \normalfont{68.018} & 47.665 & 68.708  \\  
    \bottomrule
    \bottomrule
    \multirow{4}{*}{(b)} & {Calibration}~\citep{DBLP:conf/ijcai/ShangFPFZY18} 
     & 0.5107 & 2.7129 & 5.6281 & 95.997 & 97.590 & 98.242 & 6.7281 & 10.625 & 12.322 & 0.0034 & 58.786 & 40.850 & 59.132    \\ 
     & {CVAE-GAN}~\citep{Li2019InsufficientDC}
     & 0.5175 & 2.7843 & 5.8150 & 95.303 & 97.109 & 98.218 & \textbf{6.9186} & 10.747 & 12.592 & 0.0104 & 57.610 & 40.871 & 58.767   \\ 
     & {Clustering}~\citep{csaky-etal-2019-improving}
     & 0.4728 & 2.6349 & 5.3878	& \textbf{96.145} & 97.614 & 98.317 & 6.8789 & 10.869 & 13.271 & 0.0124 & \textbf{59.069} & 41.026 & 59.343   \\ 
     \setrow{\bfseries} 
     & Ours  & 0.7285 & 3.6053 & 7.2580 & \normalfont{95.938} & 97.829 & 98.561 & \normalfont{6.8391} & 10.903 & 13.411 & 0.0210 & \normalfont{58.105} & 41.113 & 59.551 \\ 
    \bottomrule
    \end{tabular}
    }
    \caption{Performance (\%) of our approach instantiated on the naive SEQ2SEQ and the baseline approaches on (a) DailyDialog and (b) OpenSubtitles.}
    \label{tbl:compare_res}
\end{table*}

\subsection{Evaluation Metrics}
We adopt several widely used metrics~\citep{DBLP:conf/emnlp/LiuLSNCP16,DBLP:conf/naacl/LiGBGD16,DBLP:journals/corr/SerbanSLCPCB16,DBLP:conf/iclr/GuCHK19} to measure the performance of dialogue generation models, including BLEU, embedding-based metrics, entropy-based metrics and distinct metrics. 
In particular, BLEU measures how much a generated response contains n-gram overlaps with the reference. 
We compute BLEU scores for n$<$4 using smoothing techniques\footnote{\url{https://www.nltk.org/_modules/nltk/translate/bleu_score.html}}.
Embedding-based metric computes the cosine similarity of bag-of-words embeddings between the hypothesis and the reference.
We employ the following three embedding metrics to assess the response quality: 
({1}) Embedding Average (\textbf{Avg}): cosine similarity between two utterances, in which the sentence embedding is computed by taking the average word embedding weighted by the smooth inverse frequency $\textit{sent\_emb}(e)=\frac{1}{|e|}\sum_{\nu\in{}e}\frac{0.001}{0.001 + p(\nu)}emb(\nu)$ of words as in \citet{DBLP:conf/iclr/AroraLM17}.
where $emb(\nu)$ and $p(\nu)$ are the embedding and the probability\footnote{ Probability is computed based on the maximum likelihood estimation on the training data.} of word $\nu$ respectively.
({2}) Embedding Greedy (\textbf{Gre}): greedily matching words in two utterances based on the cosine similarities between their embeddings, and averaging the obtained scores,
({3}) Embedding Extrema (\textbf{Ext}): cosine similarity between the largest extreme values among the word embeddings in the two utterances.
We use Glove vectors as the word embeddings.
Regarding entropy-based metrics, we compute the n-gram entropy $\textbf{Ent-n}=-\frac{1}{|r|}\sum_{\nu\in{}r}{\log_2{p(\nu)}}$ of responses to measure their non-genericness, where the probabilities $p(\nu)$ of n-grams (n=1,2,3) are calculated based on the maximum likelihood estimation on the training data~\citep{DBLP:journals/corr/SerbanSLCPCB16}.
\textbf{Distinct} computes the diversity of the generated responses. Dist-n is defined as the ratio of unique n-grams (n=1,2,3) over all n-grams in the generated responses.
Following~\citet{DBLP:conf/iclr/GuCHK19}, we also report \textbf{Intra}-\{1,2,3\} metrics which are computed as the average of distinct values within each sampled response.

\subsection{Implementation \& Reproducibility}
For word-level dialogue augmentation, we employ the pre-trained BERT-base language model with the uncased version of tokenizer.
We follow the hyper-parameters and settings suggested in ~\citet{DBLP:conf/naacl/DevlinCLT19}. 
The replacement probability is set to 15\%. 
For back-translation in sentence-level dialogue augmentation, we use the Transformer model~\citep{DBLP:conf/nips/VaswaniSPUJGKP17} trained on En-De and En-Ru WMT'19 news translation tasks~\citep{ng2019facebook}. 
German and Russian sentences were tokenized with the Moses tokenizer~\citep{DBLP:conf/acl/KoehnHBCFBCSMZDBCH07}. 
The same hyper-parameters are used for the translation tasks, i.e., word representations of size 1024, dropout with 0.8 keep probability, feed-forward layers with dimension 4096, 6 blocks in the encoder and decoder with 16 attention heads. 
Models are optimized with Adam~\citep{DBLP:journals/corr/KingmaB14} optimizer using initial learning rate 7e-4. 
Regarding dialogue models implementation, we adopt a 2-layer bidirectional LSTM as the encoder and a unidirectional one as the decoder for both the SEQ2SEQ and CVAE.
The hidden size is set to 256, and the latent size used in CVAE is set to 64. 
The transformer model for dialogue generation is configured with 512 hidden size, 8 attention heads and 6 blocks in both the encoder and decoder.
The hyper-parameters in the baseline models are set following the original papers~\citep{Li2019InsufficientDC,DBLP:conf/ijcai/ShangFPFZY18,csaky-etal-2019-improving}.

\subsection{Evaluation Results}
To investigate the effectiveness and general applicability of the proposed framework, we instantiate our data manipulation framework on several state-of-the-art models for dialogue generation.
The automatic evaluation results of our proposed learning framework and the corresponding vanilla models are listed in Table~\ref{tbl:main_res}.
Compared with the vanilla training procedure, the proposed data manipulation framework brings solid improvements for all the three architectures regarding almost all the evaluation metrics.
Such improvements are consistent across both two conversation datasets, affirming the superiority and general applicability of our proposed framework.

We further compare our model with existing related methods.
Not surprisingly, as shown in Table~\ref{tbl:compare_res}, our data manipulation framework outperforms the baseline methods on most of metrics. 
In particular, the improvement on Distinct metrics of our model is much greater, which implies that data manipulation effectively induce the neural dialogue model generating more diverse responses.

\begin{table}[!t]
    \centering
    \resizebox{0.85\columnwidth}{!}{
    \centering
    \begin{tabular}{l>{\rowmac}c>{\rowmac}>{\rowmac}c>{\rowmac}c>{\rowmac}c<{\clearrow}}
        \toprule
         \textbf{Opponent}   & \textbf{Win} & \textbf{Loss} & \textbf{Tie} & \textbf{Kappa}  \\ \midrule
         Ours \textit{vs.} SEQ2SEQ        & 45\% & 13\% & 42\% & 0.5105  \\  
         Ours \textit{vs.} {Calibration}  & 40\% & 9\%  & 51\% & 0.4208  \\
         Ours \textit{vs.} {CVAE-GAN}     & 37\% & 14\% & 49\% & 0.4063  \\  
         Ours \textit{vs.} {Clustering}   & 41\% & 12\% & 47\% & 0.4893  \\
        \bottomrule 
    \end{tabular}
    }
    \caption{The results of human evaluation on the test set of DailyDialog.}
    \label{tbl:human_eval}
\end{table}
\begin{table*}[!th]
\centering
\resizebox{1.0\textwidth}{!}{
\begin{tabular}{>{\rowmac}l>{\rowmac}c>{\rowmac}c>{\rowmac}c>{\rowmac}c>{\rowmac}c>{\rowmac}c>{\rowmac}c>{\rowmac}c>{\rowmac}c>{\rowmac}c>{\rowmac}c>{\rowmac}c>{\rowmac}>{\rowmac}c<{\clearrow}}
\toprule
& Dist-1 & Dist-2 & Dist-3 & Intra-1 & Intra-2 & Intra-3 & Ent-1 & Ent-2 & Ent-3 & BLEU & Avg & Ext & Gre  \\
\midrule
Baseline        
& 0.8570 & 4.0123 & 7.9559 & 88.509 & 94.727 & 96.844 & 6.7783 & 10.394 & 11.719 & 0.2146 & 65.200 & 46.355 & 67.344  \\   
\quad \textit{w/ word-level augmentation}   
& 1.2205 & 6.0622 & 12.2620 & 89.916 & 95.265 & 96.627 & 6.9457 & 10.920 & 12.334 & 0.2657 & 65.315 & 46.821 & 68.025  \\ 
\quad \textit{w/ sentence-level augmentation}   
& 1.4702 & 6.7803 & 13.0910 & 91.309 & 95.772 & 97.397 & 7.0260 & 10.952 & 12.517 & 0.2721 & 66.788 & 47.464 & 67.911  \\ 
\bottomrule
\end{tabular}
}
\caption{Ablation test (\%) for word-level and sentence-level augmentations.}
\label{tbl:ablation_aug}
\end{table*}
\begin{table*}[!th]
\centering
\resizebox{1.0\textwidth}{!}{
\begin{tabular}{>{\rowmac}l>{\rowmac}c>{\rowmac}c>{\rowmac}c>{\rowmac}c>{\rowmac}c>{\rowmac}c>{\rowmac}c>{\rowmac}c>{\rowmac}c>{\rowmac}c>{\rowmac}c>{\rowmac}c>{\rowmac}>{\rowmac}c<{\clearrow}}
\toprule
& Dist-1 & Dist-2 & Dist-3 & Intra-1 & Intra-2 & Intra-3 & Ent-1 & Ent-2 & Ent-3 & BLEU & Avg & Ext & Gre \\
\midrule
Full model
& 2.0515 & 9.7186 & 18.9970 & 91.343 & 96.446 & 97.613 & {7.0858} & {11.121} & {12.545} & {0.3604} & {66.551} & 47.325 & 68.378  \\ 
\quad \textit{w/o weighting} 
& 1.8156 \down & 8.1939 \down & 15.9000 \down & 90.747 \down & 95.816 \down & 97.199 \down & 7.0976 & 11.130 & 12.731 & 0.5147 & 65.675 \down & 46.955 \down & 68.048 \down \\ 
\quad \textit{w/o augmentation} 
& 1.1456 \down & 5.4386 \down & 11.1140 \down & 86.399 \down & 92.293 \down & 94.825 \down & 6.8752 \down & 10.579 \down & 11.837 \down & 0.2002 \down & 64.937 \down & 46.540 \down & 67.541 \down \\ 
\quad \textit{w/o instance filter}
& 1.8627 \down & 8.2850 \down & 15.9400 \down & 88.551 \down & 93.445 \down & 94.419 \down & 7.1440 & 11.305 & 12.823 & 0.2813 \down & 65.606 \down & 46.912 \down & 67.863 \down \\
\bottomrule
\end{tabular}
}
\caption{Model ablation test (\%) on DailyDialog.}
\label{tbl:ablation_model}
\end{table*}

\subsection{Human Evaluation}

We use the DailyDialog as the evaluation corpus since it is more similar to our daily conversations and easier for annotators to make the judgement. 
Three graduate students are recruited to conduct manual evaluations. 
100 test messages are randomly sampled. 
We present the input messages and the corresponding responses generated by our model and the comparison model to the annotators.
The annotators are then required to compare the quality of these two responses ($\text{response}_1, \text{response}_2$), taking the following criteria into consideration: coherence, language consistency, fluency and informativeness, and evaluate among ``win'' ($\text{response}_1$ is better), ``loss'' ($\text{response}_2$ is better) and ``tie'' (they are equally good or bad).
Note that cases with different evaluation results are labeled as ``tie''.
Table~\ref{tbl:human_eval} summarizes human evaluation results.
The kappa scores indicate that the annotators came to a fair agreement in the judgement.
Compared with the baseline methods, our data manipulation approach brings about more informative and coherent replies.

\begin{figure}[!t]
\centering
  \includegraphics[width=0.99\columnwidth]{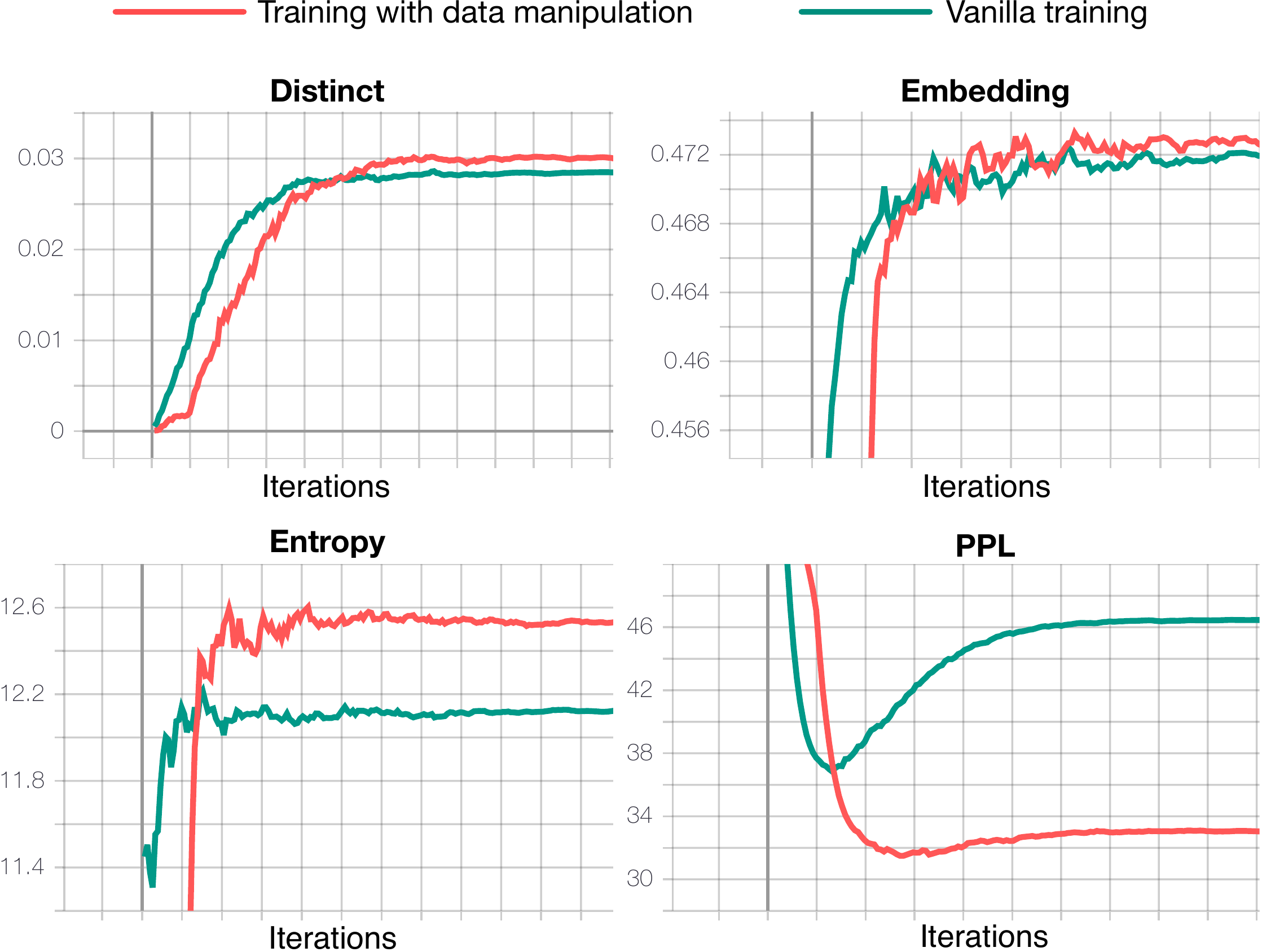}
  \caption{
  Comparison of the training with data manipulation and vanilla training using SEQ2SEQ on the validation set of DailyDialog. Dist-1, Embedding Extrema and Ent-3 are denoted as ``Distinct'', ``Embedding'' and ``Entropy'', respectively.
  }
  \label{fig:valid_line_plot}
\end{figure}

\subsection{Model Analysis}

\paragraph{Learning Efficiency}
 Figure~\ref{fig:valid_line_plot} presents validation results along iterations when training the SEQ2SEQ model on DailyDialog.
 We observe that when training SEQ2SEQ using our framework, the initial learning speed is a bit slower than the standard vanilla training.
 However, our framework surpasses the vanilla training on the final stage.
 One reason is that, at the early stage, the data manipulation model takes some time to improve its manipulation skills. 
 This may slow down the neural dialogue model learning.
 Once the manipulation skills are effective enough, the neural dialogue model may benefit from learning more effective samples instead of those inefficient instances, and achieves better performance.

\begin{figure*}[!h]
\centering
  \includegraphics[width=0.99\textwidth]{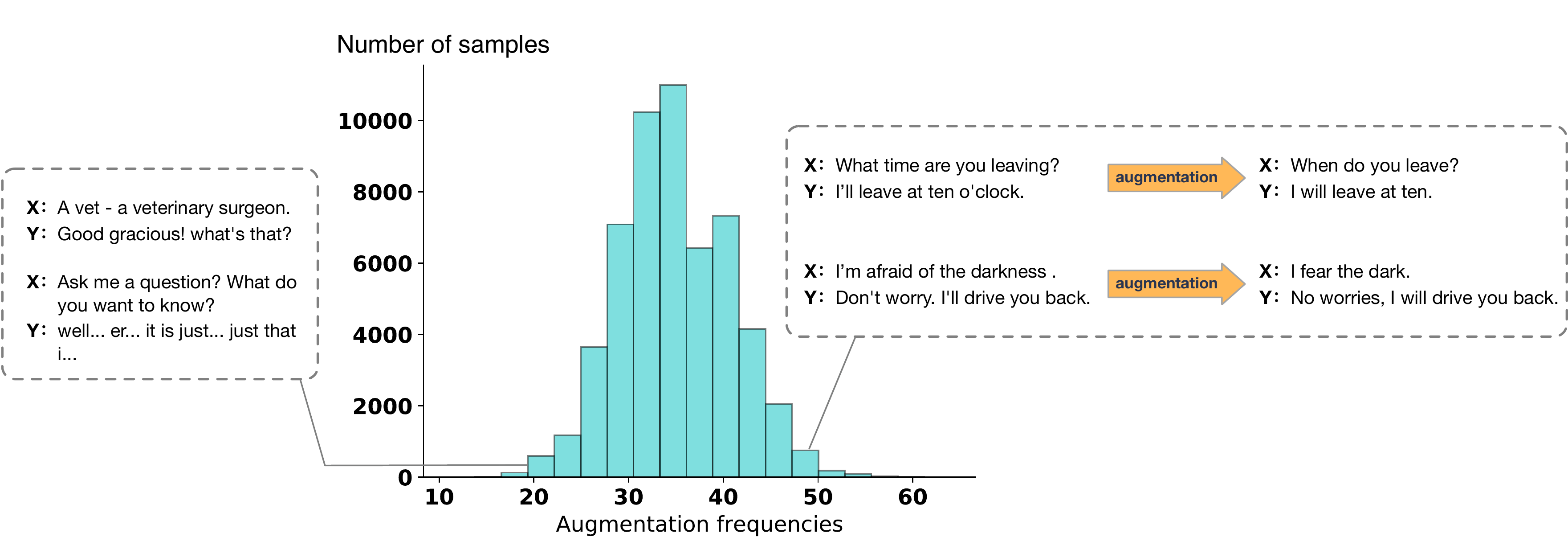}
  \caption{
    Examples with different augmentation frequencies. Instances with higher augmentation frequencies are \iffalse comprehensively\fi more effective than those seldom augmented examples.
  }
  \label{fig:gating_plot}
\end{figure*}

\paragraph{Examples with Different Augmentation Frequencies}
 The data manipulation model selectively chooses samples to conduct data augmentation.
 To further glean the insights regarding which samples are favored by the augmentation model, we list examples with different augmentation frequencies in Figure~\ref{fig:gating_plot}.
 We notice that samples frequently augmented by the manipulation model are more reliable than those seldom augmented ones.
 Therefore, the dialogue model is able to learn from those effective instances and their synonymous variants.

\paragraph{Word-level \textit{vs.} Sentence-level Augmentation}
 In our framework, we implement two kinds of augmentation mechanisms. 
 Word-level augmentation enriches the given samples by substituting words, while sentence-level augmentation paraphrases the original samples through back-translation.
 We evaluate their performances and report results in Table~\ref{tbl:ablation_aug}.
 Both augmentation mechanisms improve the performance over the vanilla SEQ2SEQ baseline,
 while sentence-level augmentation performs slightly better than word-level augmentation on most evaluation metrics.
 One possible reason is that sentence-level augmentation captures more paraphrasing phenomenon.

\paragraph{Ablation Study}
 Table~\ref{tbl:ablation_model} presents the results of model variants, by ablating specific parts of the data manipulation model.
 Among different variants, without data augmentation, the performance degrades rapidly.
 Meanwhile, without weighting or {instance filter} also decreases the performance.
 This implies that the neural dialogue generation model not only benefits from more training samples but also reaps greater advantages from those effective rather than inefficient instances.

\begin{table}[!t]
\centering

\resizebox{1.0\columnwidth}{!}{
\centering
\begin{tabular}{lcccc}
\toprule
\textbf{}                    & \textbf{Distinct} ($\bm{\Delta}$)  & \textbf{Embedding} ($\bm{\Delta}$) & \textbf{Entropy} ($\bm{\Delta}$) & \textbf{BLEU} ($\bm{\Delta}$)     \\ 
\midrule
\textbf{\makecell{50\% \\ training data}}  & \makecell{0.8179\\(+109.88\%)}  & \makecell{1.6860\\ (+2.54\%)}   & \makecell{0.4910\\ (+4.20\%)} & \makecell{0.0768\\ (+56.64\%)} \\ \hline
\textbf{\makecell{100\% \\ training data}} & \makecell{1.0865\\ (+71.62\%)}  & \makecell{0.2720\\ (+0.40\%)}   & \makecell{0.4750\\ (+3.90\%)}  & \makecell{0.1307\\ (+43.21\%)} \\
\bottomrule
\end{tabular}
}
\caption{
Performance improvements regarding different sizes of training data on DailyDialog.
Dist-1, Embedding Greedy and Ent-3 are denoted as ``Distinct'', ``Embedding'' and ``Entropy'', respectively.
}
\label{tbl:diff-training-sizes}
\end{table}

\paragraph{Impact of Training Data Scale}

We explore the impact of training data scale on the data manipulation framework by comparing a model trained on half amount of the training data in DailyDialog.
As presented in Table~\ref{tbl:diff-training-sizes}, with only 50\% amount of training data, our model achieves a greater performance boost, which affirms the effectiveness and robustness of the proposed approach.
\section{Related Work}

Existing approaches to improving neural dialogue generation models mainly target on building more powerful learning systems, using extra information such as conversation topics~\citep{DBLP:conf/aaai/XingWWLHZM17}, persona profile~\citep{DBLP:conf/ijcai/SongZCWL19}, user emotions~\citep{DBLP:conf/aaai/ZhouHZZL18}, or out-sourcing knowledge~\citep{DBLP:conf/acl/LiuFCRYL18}. 
Another popular framework for dialogue generation is variational autoencoder~\citep{DBLP:journals/corr/KingmaW13,DBLP:conf/acl/ZhaoZE17,DBLP:conf/acl/ShenSLLNZAL17}, in which a latent variable is introduced to benefit the dialogue model with more diverse response generation.
Contrasted with previous researches, we investigate to improve the dialogue model from a different angle, i.e., adapting the training examples using data manipulation techniques.

Data augmentation is an effective way to improve the performance of neural models.
To name a few,
\citet{DBLP:conf/interspeech/KurataXZ16} propose to generate more utterances by introducing noise to the decoding process. 
\citet{DBLP:conf/naacl/Kobayashi18,DBLP:conf/iccS/WuLZHH19} demonstrate that contextual augmentation using label-conditional language models helps to improve the neural networks classifier on text classification tasks.
\citet{sennrich-etal-2016-improving} boost neural machine translation models using back-translation.
\citet{DBLP:conf/iclr/XieWLLNJN17,DBLP:journals/corr/abs-1904-09545} design manually-specified strategies for data augmentation.
\citet{DBLP:conf/coling/HouLCL18} utilize a sequence-to-sequence model to produce diverse utterances for language understanding.
\citet{Li2019InsufficientDC,DBLP:journals/corr/abs-1909-12868} propose to generate sentences for dialogue augmentation.
Compared with previous augmentation approaches for dialogue generation, augmented sentences in our framework are selectively generated using the pretrained models and the augmentation process is additionally fine-tuned jointly with the training of dialogue generation. 

Regarding data weighting, past methods~\citep{jiang-zhai-2007-instance,DBLP:conf/ecml/RebbapragadaB07,DBLP:conf/emnlp/WangULCS17,DBLP:conf/icml/RenZYU18,DBLP:journals/corr/abs-1910-12795} have been proposed to manage the problem of training set biases or label noises.
\citet{lison-bibauw-2017-dialogues} propose to enhance the retrieval-based dialog system with a weighting model.
\citet{DBLP:conf/ijcai/ShangFPFZY18} likewise design a matching network to calibrate the dialogue model training through instance weighting.
\citet{cai-etal-2020-curricula} investigate curriculum learning to adapt the instance effect on dialogue model training according to the sample complexity.
Whereas our proposed framework learns to reweight not only the original training examples but also the augmented examples.
Another difference is that, we directly derive data weights based on their gradient directions on a validation set, instead of separately training a external weighting model.
~\citet{csaky-etal-2019-improving} claim that high-entropy utterances in the training set lead to those boring generated responses and thus propose to ameliorate such issue by simply removing training instances with high entropy. 
Although data filtering is a straightforward approach to alleviate the problem of noisy data, the informative training samples remain untouched and insufficient.
Whereas our method holds the promise of generating more valid training data and alleviating the negative noises in the meantime.

Note that either data augmentation or instance reweighting can be considered band-aid solution: simply augmenting all training data risks introducing more noisy conversations as such low-quality examples prevail in human-generated dialogues, whilst adapting the sample effect merely by instance reweighting is also suboptimal since effective training samples remain insufficient. The proposed learning-to-manipulate framework organically integrates these two schemes, which collectively fulfill the entire goal.

\section{Conclusion}

In this work, we consider the automated data manipulation for open-domain dialogue systems.
To induce the model learning from effective instances, we propose a learnable data manipulation model to augment effective training samples and reduce the weights of inefficient samples.
The resulting data manipulation model is fully end-to-end and can be trained jointly with the dialogue generation model.
Experiments conducted on two public conversation datasets show that our proposed framework is able to boost the performance of existing dialogue systems.

Our learning-to-manipulate framework for neural dialogue generation is not limited to the elaborately designed manipulation skills in this paper.
Future work will investigate other data manipulation techniques (e.g., data synthesis), which can be further integrated to improve the performance.

\section*{Acknowledgments}
We would like to thank all the reviewers for their insightful and valuable comments and suggestions.
This work is supported by the National Natural Science Foundation of China-Joint Fund for Basic Research of General Technology under Grant U1836111 and U1736106.
Hongshen Chen and Xiaofang Zhao are the corresponding authors.

\bibliography{Main}

\begin{thebibliography}{42}
\expandafter\ifx\csname natexlab\endcsname\relax\def\natexlab#1{#1}\fi

\bibitem[{Andreas(2020)}]{DBLP:journals/corr/abs-1904-09545}
Jacob Andreas. 2020.
\newblock Good-enough compositional data augmentation.
\newblock In \emph{ACL}.

\bibitem[{Arora et~al.(2017)Arora, Liang, and Ma}]{DBLP:conf/iclr/AroraLM17}
Sanjeev Arora, Yingyu Liang, and Tengyu Ma. 2017.
\newblock A simple but tough-to-beat baseline for sentence embeddings.
\newblock In \emph{ICLR}.

\bibitem[{Bahdanau et~al.(2015)Bahdanau, Cho, and
  Bengio}]{Bahdanau2014NeuralMT}
Dzmitry Bahdanau, Kyunghyun Cho, and Yoshua Bengio. 2015.
\newblock Neural machine translation by jointly learning to align and
  translate.
\newblock In \emph{ICLR}.

\bibitem[{Cai et~al.(2020)Cai, Chen, Zhang, Song, Zhao, Li, Duan, and
  Yin}]{cai-etal-2020-curricula}
Hengyi Cai, Hongshen Chen, Cheng Zhang, Yonghao Song, Xiaofang Zhao, Yangxi Li,
  Dongsheng Duan, and Dawei Yin. 2020.
\newblock Learning from easy to complex: Adaptive multi-curricula learning for
  neural dialogue generation.
\newblock In \emph{AAAI}.

\bibitem[{Cs{\'a}ky et~al.(2019)Cs{\'a}ky, Purgai, and
  Recski}]{csaky-etal-2019-improving}
Rich{\'a}rd Cs{\'a}ky, Patrik Purgai, and G{\'a}bor Recski. 2019.
\newblock \href {https://www.aclweb.org/anthology/P19-1567} {Improving neural
  conversational models with entropy-based data filtering}.
\newblock In \emph{ACL}.

\bibitem[{Devlin et~al.(2019)Devlin, Chang, Lee, and
  Toutanova}]{DBLP:conf/naacl/DevlinCLT19}
Jacob Devlin, Ming{-}Wei Chang, Kenton Lee, and Kristina Toutanova. 2019.
\newblock \href {https://www.aclweb.org/anthology/N19-1423} {{BERT:}
  pre-training of deep bidirectional transformers for language understanding}.
\newblock In \emph{NAACL-HLT}.

\bibitem[{Edunov et~al.(2018)Edunov, Ott, Auli, and
  Grangier}]{DBLP:conf/emnlp/EdunovOAG18}
Sergey Edunov, Myle Ott, Michael Auli, and David Grangier. 2018.
\newblock \href {https://www.aclweb.org/anthology/D18-1045} {Understanding
  back-translation at scale}.
\newblock In \emph{EMNLP}.

\bibitem[{Gu et~al.(2019)Gu, Cho, Ha, and Kim}]{DBLP:conf/iclr/GuCHK19}
Xiaodong Gu, Kyunghyun Cho, JungWoo Ha, and Sunghun Kim. 2019.
\newblock Dialogwae: Multimodal response generation with conditional
  wasserstein auto-encoder.
\newblock In \emph{ICLR}.

\bibitem[{Hou et~al.(2018)Hou, Liu, Che, and Liu}]{DBLP:conf/coling/HouLCL18}
Yutai Hou, Yijia Liu, Wanxiang Che, and Ting Liu. 2018.
\newblock \href {https://www.aclweb.org/anthology/C18-1105}
  {Sequence-to-sequence data augmentation for dialogue language understanding}.
\newblock In \emph{{COLING}}.

\bibitem[{Hu et~al.(2019)Hu, Tan, Salakhutdinov, Mitchell, and
  Xing}]{DBLP:journals/corr/abs-1910-12795}
Zhiting Hu, Bowen Tan, Ruslan Salakhutdinov, Tom~M. Mitchell, and Eric~P. Xing.
  2019.
\newblock Learning data manipulation for augmentation and weighting.
\newblock In \emph{NeurIPS}.

\bibitem[{Jang et~al.(2017)Jang, Gu, and Poole}]{DBLP:conf/iclr/JangGP17}
Eric Jang, Shixiang Gu, and Ben Poole. 2017.
\newblock Categorical reparameterization with gumbel-softmax.
\newblock In \emph{{ICLR}}.

\bibitem[{Jiang and Zhai(2007)}]{jiang-zhai-2007-instance}
Jing Jiang and ChengXiang Zhai. 2007.
\newblock \href {https://www.aclweb.org/anthology/P07-1034} {Instance weighting
  for domain adaptation in {NLP}}.
\newblock In \emph{ACL}.

\bibitem[{Kingma and Ba(2015)}]{DBLP:journals/corr/KingmaB14}
Diederik~P. Kingma and Jimmy Ba. 2015.
\newblock Adam: {A} method for stochastic optimization.
\newblock In \emph{ICLR}.

\bibitem[{Kingma and Welling(2014)}]{DBLP:journals/corr/KingmaW13}
Diederik~P. Kingma and Max Welling. 2014.
\newblock Auto-encoding variational bayes.
\newblock In \emph{ICLR}.

\bibitem[{Kobayashi(2018)}]{DBLP:conf/naacl/Kobayashi18}
Sosuke Kobayashi. 2018.
\newblock \href {https://www.aclweb.org/anthology/N18-2072} {Contextual
  augmentation: Data augmentation by words with paradigmatic relations}.
\newblock In \emph{NAACL-HLT}.

\bibitem[{Koehn et~al.(2007)Koehn, Hoang, Birch, Callison{-}Burch, Federico,
  Bertoldi, Cowan, Shen, Moran, Zens, Dyer, Bojar, Constantin, and
  Herbst}]{DBLP:conf/acl/KoehnHBCFBCSMZDBCH07}
Philipp Koehn, Hieu Hoang, Alexandra Birch, Chris Callison{-}Burch, Marcello
  Federico, Nicola Bertoldi, Brooke Cowan, Wade Shen, Christine Moran, Richard
  Zens, Chris Dyer, Ondrej Bojar, Alexandra Constantin, and Evan Herbst. 2007.
\newblock \href {https://www.aclweb.org/anthology/P07-2045} {Moses: Open source
  toolkit for statistical machine translation}.
\newblock In \emph{ACL}.

\bibitem[{Kurata et~al.(2016)Kurata, Xiang, and
  Zhou}]{DBLP:conf/interspeech/KurataXZ16}
Gakuto Kurata, Bing Xiang, and Bowen Zhou. 2016.
\newblock Labeled data generation with encoder-decoder {LSTM} for semantic slot
  filling.
\newblock In \emph{INTERSPEECH}.

\bibitem[{Li et~al.(2016)Li, Galley, Brockett, Gao, and
  Dolan}]{DBLP:conf/naacl/LiGBGD16}
Jiwei Li, Michel Galley, Chris Brockett, Jianfeng Gao, and Bill Dolan. 2016.
\newblock \href {https://www.aclweb.org/anthology/N16-1014} {A
  diversity-promoting objective function for neural conversation models}.
\newblock In \emph{NAACL-HLT}.

\bibitem[{Li et~al.(2019)Li, Qiu, Tang, Chen, Zhao, and
  Yan}]{Li2019InsufficientDC}
Juntao Li, Lisong Qiu, Bo~Tang, Min~Dong Chen, Dongyan Zhao, and Rui Yan. 2019.
\newblock Insufficient data can also rock! learning to converse using smaller
  data with augmentation.
\newblock In \emph{AAAI}.

\bibitem[{Li et~al.(2017)Li, Su, Shen, Li, Cao, and
  Niu}]{DBLP:conf/ijcnlp/LiSSLCN17}
Yanran Li, Hui Su, Xiaoyu Shen, Wenjie Li, Ziqiang Cao, and Shuzi Niu. 2017.
\newblock \href {https://www.aclweb.org/anthology/I17-1099} {Dailydialog: {A}
  manually labelled multi-turn dialogue dataset}.
\newblock In \emph{IJCNLP}.

\bibitem[{Lison and Bibauw(2017)}]{lison-bibauw-2017-dialogues}
Pierre Lison and Serge Bibauw. 2017.
\newblock \href {https://www.aclweb.org/anthology/W17-5546} {Not all dialogues
  are created equal: Instance weighting for neural conversational models}.
\newblock In \emph{SIGDIAL}.

\bibitem[{Lison and Tiedemann(2016)}]{DBLP:conf/lrec/LisonT16}
Pierre Lison and J{\"{o}}rg Tiedemann. 2016.
\newblock \href {https://www.aclweb.org/anthology/L16-1147} {Opensubtitles2016:
  Extracting large parallel corpora from movie and {TV} subtitles}.
\newblock In \emph{LREC}.

\bibitem[{Liu et~al.(2016)Liu, Lowe, Serban, Noseworthy, Charlin, and
  Pineau}]{DBLP:conf/emnlp/LiuLSNCP16}
Chia{-}Wei Liu, Ryan Lowe, Iulian Serban, Michael Noseworthy, Laurent Charlin,
  and Joelle Pineau. 2016.
\newblock \href {https://www.aclweb.org/anthology/D16-1230} {How {NOT} to
  evaluate your dialogue system: An empirical study of unsupervised evaluation
  metrics for dialogue response generation}.
\newblock In \emph{EMNLP}.

\bibitem[{Liu et~al.(2018)Liu, Chen, Ren, Feng, Liu, and
  Yin}]{DBLP:conf/acl/LiuFCRYL18}
Shuman Liu, Hongshen Chen, Zhaochun Ren, Yang Feng, Qun Liu, and Dawei Yin.
  2018.
\newblock \href {https://www.aclweb.org/anthology/P18-1138} {Knowledge
  diffusion for neural dialogue generation}.
\newblock In \emph{ACL}.

\bibitem[{Ng et~al.(2019)Ng, Yee, Baevski, Ott, Auli, and
  Edunov}]{ng2019facebook}
Nathan Ng, Kyra Yee, Alexei Baevski, Myle Ott, Michael Auli, and Sergey Edunov.
  2019.
\newblock \href {https://www.aclweb.org/anthology/W19-5333} {Facebook fair's
  wmt19 news translation task submission}.
\newblock In \emph{Proc. of WMT}.

\bibitem[{Niu and Bansal(2019)}]{DBLP:journals/corr/abs-1909-12868}
Tong Niu and Mohit Bansal. 2019.
\newblock \href {https://www.aclweb.org/anthology/D19-1132} {Automatically
  learning data augmentation policies for dialogue tasks}.
\newblock In \emph{EMNLP}.

\bibitem[{Rebbapragada and Brodley(2007)}]{DBLP:conf/ecml/RebbapragadaB07}
Umaa Rebbapragada and Carla~E. Brodley. 2007.
\newblock Class noise mitigation through instance weighting.
\newblock In \emph{{ECML}}.

\bibitem[{Ren et~al.(2018)Ren, Zeng, Yang, and
  Urtasun}]{DBLP:conf/icml/RenZYU18}
Mengye Ren, Wenyuan Zeng, Bin Yang, and Raquel Urtasun. 2018.
\newblock Learning to reweight examples for robust deep learning.
\newblock In \emph{ICML}.

\bibitem[{Sennrich et~al.(2016)Sennrich, Haddow, and
  Birch}]{sennrich-etal-2016-improving}
Rico Sennrich, Barry Haddow, and Alexandra Birch. 2016.
\newblock \href {https://www.aclweb.org/anthology/P16-1009} {Improving neural
  machine translation models with monolingual data}.
\newblock In \emph{ACL}.

\bibitem[{Serban et~al.(2017)Serban, Sordoni, Lowe, Charlin, Pineau, Courville,
  and Bengio}]{DBLP:journals/corr/SerbanSLCPCB16}
Iulian~Vlad Serban, Alessandro Sordoni, Ryan Lowe, Laurent Charlin, Joelle
  Pineau, Aaron~C. Courville, and Yoshua Bengio. 2017.
\newblock A hierarchical latent variable encoder-decoder model for generating
  dialogues.
\newblock In \emph{AAAI}.

\bibitem[{Shang et~al.(2018)Shang, Fu, Peng, Feng, Zhao, and
  Yan}]{DBLP:conf/ijcai/ShangFPFZY18}
Mingyue Shang, Zhenxin Fu, Nanyun Peng, Yansong Feng, Dongyan Zhao, and Rui
  Yan. 2018.
\newblock Learning to converse with noisy data: Generation with calibration.
\newblock In \emph{{IJCAI}}.

\bibitem[{Shen et~al.(2017)Shen, Su, Li, Li, Niu, Zhao, Aizawa, and
  Long}]{DBLP:conf/acl/ShenSLLNZAL17}
Xiaoyu Shen, Hui Su, Yanran Li, Wenjie Li, Shuzi Niu, Yang Zhao, Akiko Aizawa,
  and Guoping Long. 2017.
\newblock \href {https://aclweb.org/anthology/P17-2080} {A conditional
  variational framework for dialog generation}.
\newblock In \emph{ACL}.

\bibitem[{Song et~al.(2019)Song, Zhang, Cui, Wang, and
  Liu}]{DBLP:conf/ijcai/SongZCWL19}
Haoyu Song, Weinan Zhang, Yiming Cui, Dong Wang, and Ting Liu. 2019.
\newblock Exploiting persona information for diverse generation of
  conversational responses.
\newblock In \emph{IJCAI}.

\bibitem[{Tan et~al.(2019)Tan, Hu, Yang, Salakhutdinov, and
  Xing}]{DBLP:conf/iclr/TanHYSX19}
Bowen Tan, Zhiting Hu, Zichao Yang, Ruslan Salakhutdinov, and Eric~P. Xing.
  2019.
\newblock Connecting the dots between {MLE} and {RL} for sequence generation.
\newblock In \emph{{ICLR} Workshop}.

\bibitem[{Vaswani et~al.(2017)Vaswani, Shazeer, Parmar, Uszkoreit, Jones,
  Gomez, Kaiser, and Polosukhin}]{DBLP:conf/nips/VaswaniSPUJGKP17}
Ashish Vaswani, Noam Shazeer, Niki Parmar, Jakob Uszkoreit, Llion Jones,
  Aidan~N. Gomez, Lukasz Kaiser, and Illia Polosukhin. 2017.
\newblock Attention is all you need.
\newblock In \emph{NIPS}.

\bibitem[{Wang et~al.(2017)Wang, Utiyama, Liu, Chen, and
  Sumita}]{DBLP:conf/emnlp/WangULCS17}
Rui Wang, Masao Utiyama, Lemao Liu, Kehai Chen, and Eiichiro Sumita. 2017.
\newblock \href {https://www.aclweb.org/anthology/D17-1155} {Instance weighting
  for neural machine translation domain adaptation}.
\newblock In \emph{{EMNLP}}.

\bibitem[{Wu et~al.(2019)Wu, Lv, Zang, Han, and Hu}]{DBLP:conf/iccS/WuLZHH19}
Xing Wu, Shangwen Lv, Liangjun Zang, Jizhong Han, and Songlin Hu. 2019.
\newblock Conditional {BERT} contextual augmentation.
\newblock In \emph{{ICCS}}.

\bibitem[{Xie et~al.(2017)Xie, Wang, Li, L{\'{e}}vy, Nie, Jurafsky, and
  Ng}]{DBLP:conf/iclr/XieWLLNJN17}
Ziang Xie, Sida~I. Wang, Jiwei Li, Daniel L{\'{e}}vy, Aiming Nie, Dan Jurafsky,
  and Andrew~Y. Ng. 2017.
\newblock Data noising as smoothing in neural network language models.
\newblock In \emph{{ICLR}}.

\bibitem[{Xing et~al.(2017)Xing, Wu, Wu, Liu, Huang, Zhou, and
  Ma}]{DBLP:conf/aaai/XingWWLHZM17}
Chen Xing, Wei Wu, Yu~Wu, Jie Liu, Yalou Huang, Ming Zhou, and Wei{-}Ying Ma.
  2017.
\newblock Topic aware neural response generation.
\newblock In \emph{AAAI}.

\bibitem[{Yu et~al.(2018)Yu, Dohan, Luong, Zhao, Chen, Norouzi, and
  Le}]{DBLP:conf/iclr/YuDLZ00L18}
Adams~Wei Yu, David Dohan, Minh{-}Thang Luong, Rui Zhao, Kai Chen, Mohammad
  Norouzi, and Quoc~V. Le. 2018.
\newblock Qanet: Combining local convolution with global self-attention for
  reading comprehension.
\newblock In \emph{ICLR}.

\bibitem[{Zhao et~al.(2017)Zhao, Zhao, and
  Esk{\'{e}}nazi}]{DBLP:conf/acl/ZhaoZE17}
Tiancheng Zhao, Ran Zhao, and Maxine Esk{\'{e}}nazi. 2017.
\newblock \href {https://www.aclweb.org/anthology/P17-1061/} {Learning
  discourse-level diversity for neural dialog models using conditional
  variational autoencoders}.
\newblock In \emph{ACL}.

\bibitem[{Zhou et~al.(2018)Zhou, Huang, Zhang, Zhu, and
  Liu}]{DBLP:conf/aaai/ZhouHZZL18}
Hao Zhou, Minlie Huang, Tianyang Zhang, Xiaoyan Zhu, and Bing Liu. 2018.
\newblock Emotional chatting machine: Emotional conversation generation with
  internal and external memory.
\newblock In \emph{AAAI}.

\end{thebibliography}
\bibliographystyle{ACL/acl_natbib}

\end{document}